\documentclass{article}
\usepackage{ijcai24}

\usepackage{times}
\usepackage{soul}
\usepackage{url}
\usepackage[hidelinks]{hyperref}
\usepackage[utf8]{inputenc}
\usepackage[small]{caption}
\usepackage{graphicx}
\usepackage{amsmath}
\usepackage{amsthm}
\usepackage{booktabs}
\usepackage{algorithm}
\usepackage{algorithmic}
\usepackage[switch]{lineno}
\usepackage{amsfonts,amssymb}
\usepackage[utf8]{inputenc}
\usepackage{url}
\usepackage{booktabs}
\usepackage{amssymb}
\usepackage{bbding}
\usepackage{pifont}
\usepackage{wasysym}
\usepackage{utfsym}
\usepackage{fontawesome}
\usepackage{multirow}

\usepackage[utf8]{inputenc}
\usepackage{url}
\usepackage{booktabs}
\usepackage{amssymb}
\usepackage{bbding}
\usepackage{pifont}
\usepackage{wasysym}
\usepackage{utfsym}
\usepackage{fontawesome}

\title{Multimodal Task Representation Memory Bank vs. Catastrophic Forgetting in Anomaly Detection }
\author{
    You Zhou, Jiangshan Zhao, Deyu Zeng, Zuo Zuo,Weixiang Liu, Zongze Wu
}

\begin{document}

\maketitle

\begin{abstract}
Unsupervised Continuous Anomaly Detection (UCAD) faces significant challenges in multi-task representation learning, with existing methods suffering from incomplete representation and catastrophic forgetting. Unlike supervised models, unsupervised scenarios lack prior information, making it difficult to effectively distinguish redundant and complementary multimodal features.
To address this, we propose the Multimodal Task Representation Memory Bank (MTRMB) method through two key technical innovations: A Key-Prompt-Multimodal Knowledge (KPMK) mechanism that uses concise key prompts to guide cross-modal feature interaction between BERT and ViT.
Refined Structure-based Contrastive Learning (RSCL) leveraging Grounding DINO and SAM to generate precise segmentation masks, pulling features of the same structural region closer while pushing different structural regions apart.
Experiments on MVtec AD and VisA datasets demonstrate MTRMB's superiority, achieving an average detection accuracy of 0.921 at the lowest forgetting rate, significantly outperforming state-of-the-art methods. We plan to open source on GitHub.
\end{abstract}





\section{Introduction}

\begin{figure}[t]
	\includegraphics[width=1\columnwidth, trim=0 0 0 0, clip]{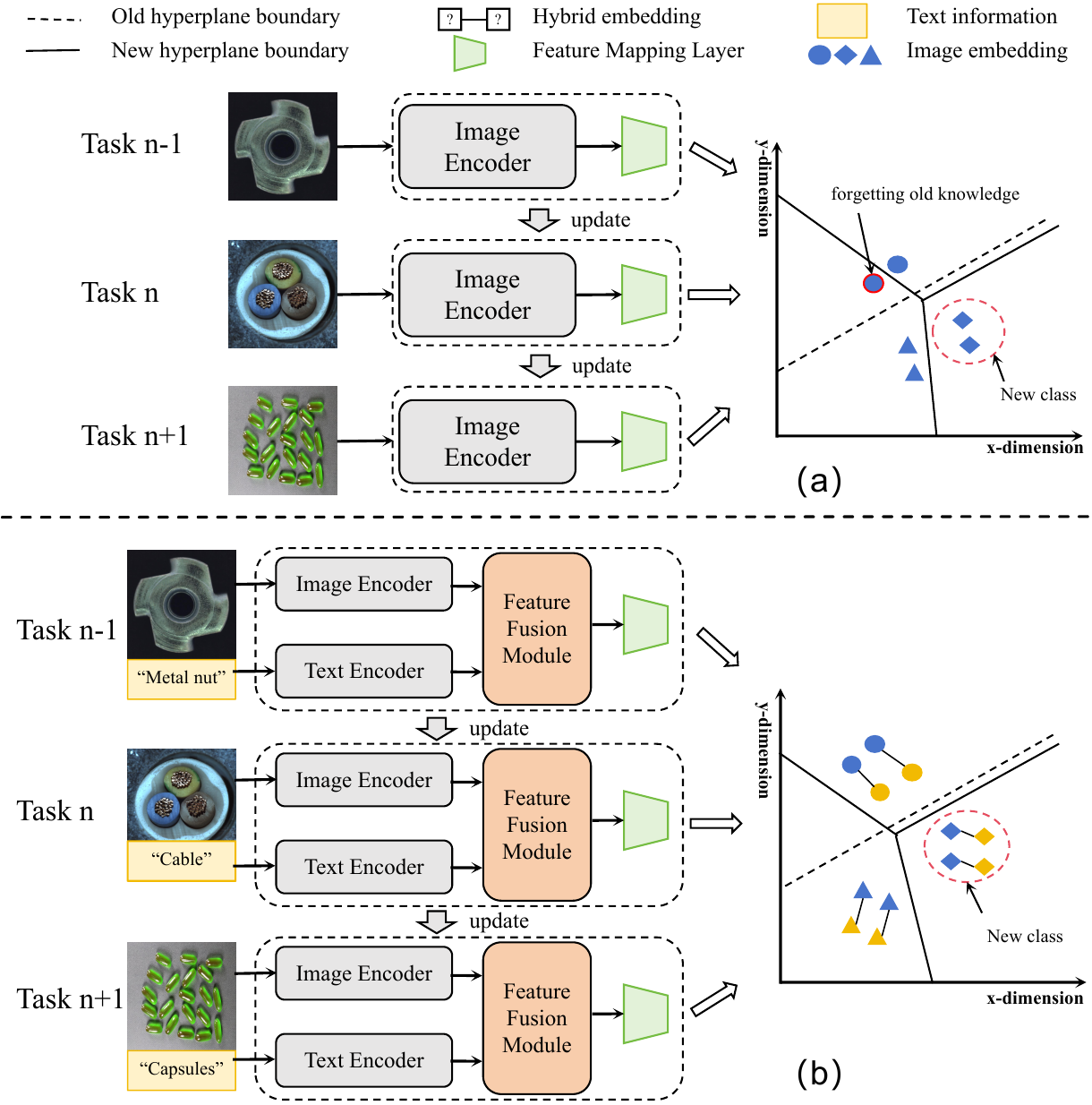} 
	\caption{Comparison between UCAD with a memory bank and UCAD with a multimodal Task Representation Memory Bank. a) The memory bank-based method has the problem of forgetting old task knowledge due to incomplete representation, resulting in inaccurate new task hyperplane boundary. b) Introducing multimodal features to construct a task representation memory can effectively retain the intrinsic information of the old task and obtain a better new task hyperplane boundary.}
	\label{fig1} 
\end{figure}


Anomaly detection (AD) \cite{lin2024survey} refers to the identification and location of anomalies when there is limited or no prior knowledge of the anomaly. In recent years, the industry has begun to focus on equipping AD with continuous learning capabilities, that is, continuously learning the intrinsic distribution of normal data in new tasks without forgetting the knowledge of old tasks. This method is very necessary in industrial inspection scenarios because it is very difficult and expensive to collect defect data in new scenarios.


Recently, most AD \cite{lin2024survey} focuses on training specific models for detection, which is effective for a single task. However, this suffers from a very serious catastrophic forgetting problem \cite{zhou2024class}. In addition, a single AD model is difficult to migrate, resulting in a waste of computing resources and an inability to address the privacy issues of industrial data. To support anomaly detection for multiple tasks, some AD methods \cite{you2022unified,yao2024prior,ho2024long} aim to train a unified model for multi-class anomaly detection. In actual industrial scenarios, the difficulty in providing training data for all tasks and the heavy computational burden hinder the practical application of such methods. In addition, such methods do not solve the catastrophic forgetting problem from the root.



With the development of AD, a few researchers began to realize the importance of Unsupervised Continuous Anomaly Detection(UCAD). DNE \cite{li2022towards} introduced continuous learning into the anomaly detection task for the first time, alleviating the catastrophic forgetting problem in the training phase of the continuous anomaly detection model. However, DNE can only be used to detect anomalies but not to locate abnormal areas. As shown in Figure \ref{fig1} (a), UCAD based on image feature memory \cite{liu2024unsupervised} solves the problem that DNE cannot locate abnormal areas, but there is a catastrophic forgetting problem caused by incomplete representation.

The introduction of multimodal prompts can improve catastrophic forgetting caused by incomplete representation. By applying a small number of prompt parameters in a continuous space to modify the input, it can rely on the learnable prompts \cite{wang2022learning} of the language and visual encoders to achieve continuous learning across tasks and prevent forgetting. However, Prompt-based multimodal continuous learning (MMCL) \cite{yu2024recent} applied to unsupervised anomaly detection cannot improve the multimodal perception ability of the model through prior information like the supervised model. In the absence of supervised information, learnable prompts cannot be used to discriminate the redundant and complementary information of multimodal features, thereby achieving continuous learning in anomaly detection. Therefore, it is crucial to explore the application of multimodal prompt incremental learning in UCAD.

To address the above problems, the proposed multimodal Task Representation Memory Bank (MTRMB) can solve catastrophic forgetting in anomaly detection. MTRMB utilizes a memory space of Key-prompts-Multimodal knowledge (KPMK) to support Continuous Learning in AD, as shown in Figure \ref{fig1} (b). In addition, Refined Structure-based Contrastive Learning (RSCL) leveraging Grounding DINO \cite{liu2025grounding} and Segment Anything Model (SAM) \cite{kirillov2023segment} to generate precise segmentation masks, pulling features of the same structural region closer while pushing different structural regions apart, and make MTRMB becomes more compact. During the training phase, the proposed memory space of Key-prompts-Multimodal knowledge stores the key, prompts, and multimodal knowledge of a specific task. During the testing phase, MMCL implements task matching, task domain adaptation, and the transfer of "normal" knowledge of different classes. Given a test image, MTRMB will automatically query the task key to retrieve the corresponding task prompt, complete the model's adaptation to the task domain through the prompt, and then extract image features and perform similarity calculation with normal knowledge, similar to PatchCore \cite{li2022towards}. However, the frozen backbone (ViT) cannot provide compact feature representations across different tasks. To overcome this problem, RSCL using accurate masks from the Grounding DINO and SAM is proposed to acquire a more compact MTRMB. We obtain more representative contextual features between different classes through structural contrastive learning, where features of the same structure are pulled together and pushed away from features of other structures, which can effectively reduce domain shift \cite{kirillov2023segment}. We conduct extensive experiments and the results demonstrate the advancedness of the proposed MTRMB. Our contributions can be summarized as follows:

\begin{itemize}
\item To the best of our knowledge, the proposed MTRMB is the first to use incremental learning of multimodal prompts for unsupervised anomaly detection. MTRMB proposes a Key-Prompt-Multimodal Knowledge mechanism for task matching, knowledge transfer, unsupervised anomaly detection, and segmentation.

\item The proposed RSCL to obtain more compact MTRMB by leveraging accurate masks from Grounding DINO and SAM. Specifically, RSCL pulls features of the same structural region closer and pushes features of different structural regions farther apart according to the multimodal model interaction.

\item We have conducted thorough experiments and introduced a new benchmark for UCAD. Compared with other advanced methods on MVTec AD and VisA, MTRMB showed superior performance, with an average detection accuracy of 0.921 under the lowest forgetting rate.
\end{itemize}

\section{Related Work}
\subsection{Anomaly Detection}
With the release of the MVTec AD dataset \cite{bergmann2019mvtec} and Visa \cite{zou2022spot}, the development of industrial image anomaly detection has shifted from a supervised paradigm to an unsupervised paradigm. research on common AD \cite{liu2024deep} has been divided into two main categories: feature-embedding-based methods and reconstruction-based methods.
\textbf{Feature-embedding-based methods} can be further categorized into four subcategories, including teacher-student model\cite{bergmann2020uninformed,salehi2021multiresolution,deng2022anomaly,tien2023revisiting,batzner2024efficientad}, one-class classification methods \cite{liu2023simplenet,cao2023anomaly}, mapping-based methods \cite{zhou2024msflow,rudolph2022fully} and memory-based methods \cite{li2022towards,xie2023pushing}. \textbf{Reconstruction-based
methods} can be further divided into Autoencoders-based \cite{schluter2022natural,zavrtanik2021draem}, Generative Adversarial Networks-based \cite{peng2024industrial}.

However, existing AD methods train separate anomaly models for different classes, which inevitably suffers from catastrophic forgetting and excessive computational burden. Even the multi-class unified anomaly detection model \cite{you2022unified,zhao2023omnial,ho2024long} does not consider the case of continuous anomaly detection. Our method is specifically designed for the scenario of continuous learning and achieves continuous anomaly detection and segmentation in an unsupervised manner.

\begin{figure*}[h]
	\centering 
	\includegraphics[scale=0.32]{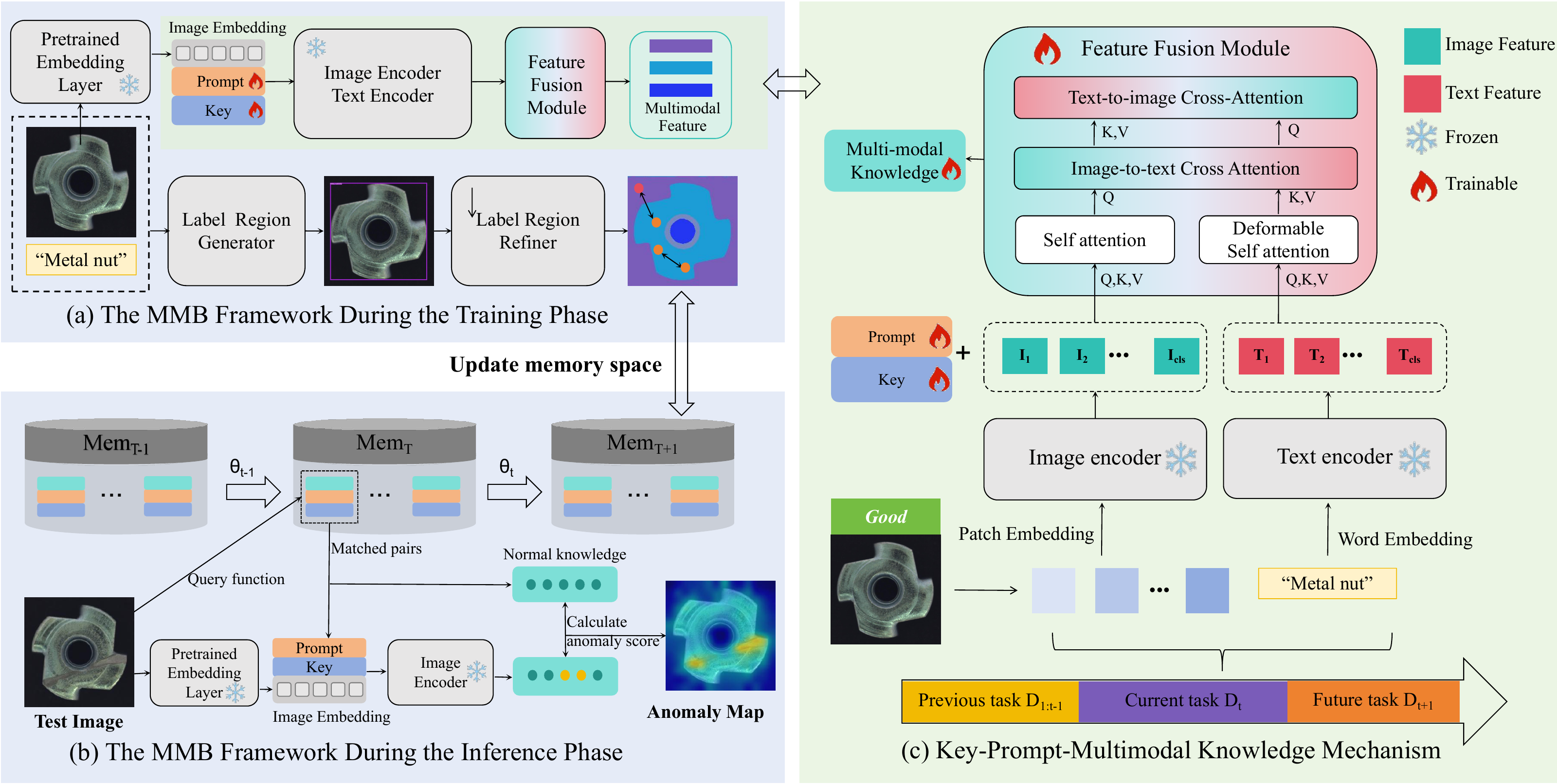} 
	\caption{The framework of UCAD using multimodal Task Representation Memory Bank.
    (a) Text-image data is input during the training phase, and an effective task intrinsic memory bank is formed through the KPMK. In addition, we use RSCL to better utilize task-related contextual information to obtain a more compact MTRMB. (b) When a test image is input during the testing phase, the framework automatically queries the Task key to retrieve the corresponding task prompts, completes the model's transfer of task knowledge through the prompts, then extracts the features of the test image and calculates the similarity with normal knowledge, and finally completes continuous detection of anomalies. (c) The KPMK mechanism uses the concise key to guide the cross-fusion of features from two different modalities, text and image, and generates an effective task representation memory bank.}
	\label{MTRMB} 
\end{figure*}


\subsection{Continual Anomaly Detection}
IDDM \cite{zhang2023iddm} proposed an incremental anomaly detection method based on a small number of labeled samples. On the other hand, LeMO \cite{li2023cross} follows the common unsupervised anomaly detection paradigm and performs incremental anomaly detection as normal samples continue to increase. However, both IDDM and LeMO focus on the study of intra-class continuous anomaly detection, but do not address the challenge of inter-class incremental anomaly detection. Li et al. \cite{li2022towards} proposed DNE for image-level anomaly detection in continuous learning scenarios. Due to the limitation that DNE only stores class-level information, it cannot perform fine-grained localization and is therefore not suitable for anomaly segmentation. UCAD solves the problem that DNE cannot locate abnormal areas and further alleviates the problems of catastrophic forgetting and excessive computational burden. However, this method suffers from performance limitations due to the lack of compactness and comprehensiveness of the representation.


\section{Methods}

\subsection{Key-Prompt-Multimodal Knowledge Mechanism}
Using MTRMB to solve the problem of catastrophic forgetting in continuous learning faces three major problems: 1) Existing unsupervised anomaly detection methods have serious catastrophic forgetting problems. 2) Single-modal feature representations are difficult to fully retain the intrinsic information of old tasks. 3) It is costly to collect abnormal data for different tasks in industrial scenarios, and an effective continuous learning mechanism is required. To solve the above problems, we propose a key prompt-multimodal knowledge (KPMK) mechanism, which can effectively fuse text (BERT) and image (ViT) cross-modal features to build a memory bank that can retain the essential knowledge of the task. This structure only stores positive sample knowledge that takes up less memory and lightweight prompt vectors and keys, which can effectively complete task key queries, task adaptation, knowledge transfer, and complete incremental anomaly detection for different tasks.


In the task key query stage, we selected the feature vector extracted from the fifth layer of the pretrained vision transformer (ViT) on normal data as the task key. This is because the features of the middle layer often contain rich contextual information, which can better represent the task itself. In order to obtain a more effective key, we used the FPS method to further condense the key set of all tasks.


In the task adaptation phase, we adopt Prefix Tuning to insert learnable prompts $P \in \mathbb{R}^{L_p \times D}$ into different layers of a pre-trained ViT. where, $L_p$ denotes the prompt length, and $D$ represents the embedding dimension. Given an input image $X \in \mathbb{R}^{H \times W \times C}$, it is processed through ViT $f$ to obtain the feature embedding $ X_e \in \mathbb{R}^{L \times D} $. The learnable prompt is then concatenated with $X_e$, forming the final input:

\begin{equation}
    f(X_e, P) = \text{MSA}(W_i^Q X_e, [P_k; W_i^K X_e], [P_v; W_i^V X_e])
\end{equation}

where $ W_i^Q, W_i^K, W_i^V $ are projection matrices.

Furthermore, each prompt $ P_i $ is associated with a learnable key, represented as $ (K_i, P_i) $, where $ K_i \in \mathbb{R}^{D} $. A query function $ q(x) $ maps the input features to the key space, and the best-matching key is selected based on cosine similarity:

\begin{equation}
    L_{kp} = \arg\min \sum_{i=1}^{N} \gamma(q(x), K_i)
\end{equation}

where $ \gamma $ denotes cosine similarity, and $ N $ is the number of keys.

In the knowledge transfer stage, To obtain a more comprehensive representation, we fuse text and image information through a cross-modal attention mechanism, as shown in Figure \ref{MTRMB} (c).  Specifically, Text-to-Image Attention employs image features $I$ as query $Q$ and text features $T$ as the key $K$ and value $V$, enabling the model can enhance the image representations based on textual information,thereby incorporating key content described in the text into the image features. Image-to-Text Attention utilizes text features $T$ as the query $Q$ and image features $I$ as the key $K$ and value $V$. By computing the similarity between text and image features, the model refines the textual representations based on the image features, leading to more accurate text descriptions.The formula is as follows:

\begin{equation}
\begin{aligned}
&Q=IW_Q,\quad K=TW_K,\quad V=TW_V\\
&Attention(Q,K,V)=softmax(QK^T/\sqrt{D_K})
\end{aligned}
\end{equation}

\begin{equation}
\begin{aligned}
&Q=TW_Q,\quad K=IW_K,\quad V=IW_V\\
&Attention(Q,K,V)=softmax(QK^T/\sqrt{D_K})
\end{aligned}
\end{equation}


Finally, a task-related $\{K,P,MK_n\}$ is formed for continuous anomaly detection, which transferring knowledge from the previous task to the current image. However, the features stored in $MK_n$ may not be discriminative enough because the backbone has been pre-trained and is not adapted to the current task. To make the feature representation more compact, we develop RSCL for fast contrastive learning.

\subsection{Refined Structure-based Contrastive Learning}
Inspired by ReConPatch \cite{hyun2024reconpatch}, we propose a refined structure-based contrastive learning method aimed at enhancing network representations for patch-level comparisons during testing, as shown in Figure \ref{MTRMB} (a). Our approach leverages Grounding Dino and SAM to consistently provide general structure knowledge, such as segmentation masks, without requiring additional training. 
As illustrated in Figure \ref{fig2}, for each image in the training set, we employed Grounding Dino and SAM to generate segmentation masks, in which different regions represent distinct structures or semantics. Simultaneously, guided by prompts, We use $f(P,X_e)\in \mathbb{R}^{N \times M \times C}$ to represent the patch-level features extracted by ViT, where $N$ represents the batch size, $M$ represents the number of patch features of each sample, and $C$ represents the dimension of each patch feature. Grounding dino and SAM are utilized to segment input images and generate pseudo labels $Y\in \mathbb{R}^{N \times H \times W}$, where $N$ represents the batch size, and $H$ and $W$ represent the height and width of each pseudo label, respectively.


The loss function is:

\begin{equation}
    S_{i,j} = \frac{\cos(f_i(P,X_e), f_j(P,X_e))}{\tau},\quad
        M_{i,j} =
    \begin{cases}
      1 & \text{if } Y_i = Y_j \\
      0 & \text{else}
    \end{cases}
\end{equation}

The region-level contrast loss is as follows:

\begin{equation}
    L_{\text{costra}} = \frac{1}{M^2} \sum_{i=1}^{M} \sum_{j=1}^{M} \left( -S_{i,j} * M_{i,j} + (1 - M_{i,j}) * e^{S_{i,j}} \right)
\end{equation}

where $S \in \mathbb{R}^{M \times M}$ represents the similarity matrix between features, and $\tau$ is the temperature coefficient,which represents the similarity between the $i$-th eigenvector and the $j$-th eigenvector, and the similarity measurement uses cosine similarity.$f_i(P,X_e), f_j(P,X_e)$ respectively represent the representation of the $i$-th patch and the $j$-th patch of the current feature map. $M\in\mathbb{R}^{M \times M}$ is a binary mask matrix. $Y_i$ represents the label of the $i$-th patch. When the labels of two patches are the same, the value of the mask matrix is 1, indicating a positive sample pair; when the labels are different, the value of the mask matrix is 0, indicating a negative sample pair.

The cross-modal contrast loss is as follows:

\begin{equation}
\begin{aligned}
    L_{\text{cross}} = - \frac{1}{N \times M} \sum_{i=1}^{N} \sum_{j=1}^{M} \log \left( \frac{e^{\frac{\cos(f_{i,j}(P,X_e), T_{i,j})}{\tau}}}{\sum_{k=1}^{M} e^{\frac{\cos(f_{i,k}(P,X_e), T_{i,k})}{\tau}}} \right)
\end{aligned}
\end{equation}

where $T\in \mathbb{R}^{N\times 1\times D}$ denotes features extracted by the text encoder,$N$ represents the batch size and $D$ is the dimensionality of the text feature.

The Total loss is as follows:

\begin{equation}
    L_{\text{all}} = L_{\text{contra}} + L_{\text{cross}}+ \lambda L_{kp}
\end{equation}

\subsection{Inference Process}

As shown in Figure \ref{MTRMB} (b), during the inference, MTRMB retrieves stored knowledge for anomaly detection. For a test image $x$, MTRMB first finds a relevant prompt $P$ using a key-value query mechanism. The prompt $P$ is then concatenated with the embedded feature $X_e$ of the image to form a patch-level feature map, guiding the model to extract features more relevant to the task. By calculating the similarity between this feature map and the image knowledge $K_I$ stored in the memory bank, the system identifies the most relevant knowledge $k_I$ corresponding to the current task. Once the relevant knowledge is obtained, the system calculates the maximum and minimum values of the distances between the patch features of the current test image and those of the stored normal images, thereby constructing an anomaly score.

The related formulas are as follows:
\begin{equation}
\begin{aligned}
    MK_{t} = \mathop{\arg \min}_{k_I \in K_I \in MK}\sum _{i=1}^{N}\min _{j \in N}||V_o([X_e, P])-K_{I}||_{2}
\end{aligned}
\end{equation}

\begin{equation}
\begin{aligned}
F^{test}, F=\mathop{\arg \max}_{F^{test}\in V_o([X_e, P])}\mathop{\arg\min }_{F\in K_{n}^{t}}||F^{test}-F||_{2}
\end{aligned}
\end{equation}

\begin{equation}
\begin{aligned}
 S=||F^{test}-F||_{2} 
\end{aligned}
\end{equation}

where $V_o$ represents the patch-level feature extracted by the pre-trained VIT network at layer $o$ after concatenating the embedded feature of the test image with the relevant prompt $P$. Here, $o=5$. $F^{test}$ denotes the patch-level feature extracted by the VIT network at layer $o$ from the test image, and $F$ represents the corresponding patch feature of the stored normal image. By finding the minimum distance between the same patch features, the matched test and stored patches are obtained. Then, the Euclidean distance between these corresponding patch features is calculated to derive their anomaly score.

\section{Experiments and Discussion}\label{exp_dis}
\subsection{Experiments Setup}
\textbf{Datasets:} MVTec AD \cite{bergmann2019mvtec} is the most widely used dataset for industrial image anomaly detection. VisA \cite{zou2022spot} is now the largest dataset for real-world industrial anomaly detection with pixel-level annotations. We conduct experiments on these two datasets.
\begin{figure}[t]
	\includegraphics[width=1\columnwidth, trim=0 0 0 0, clip]{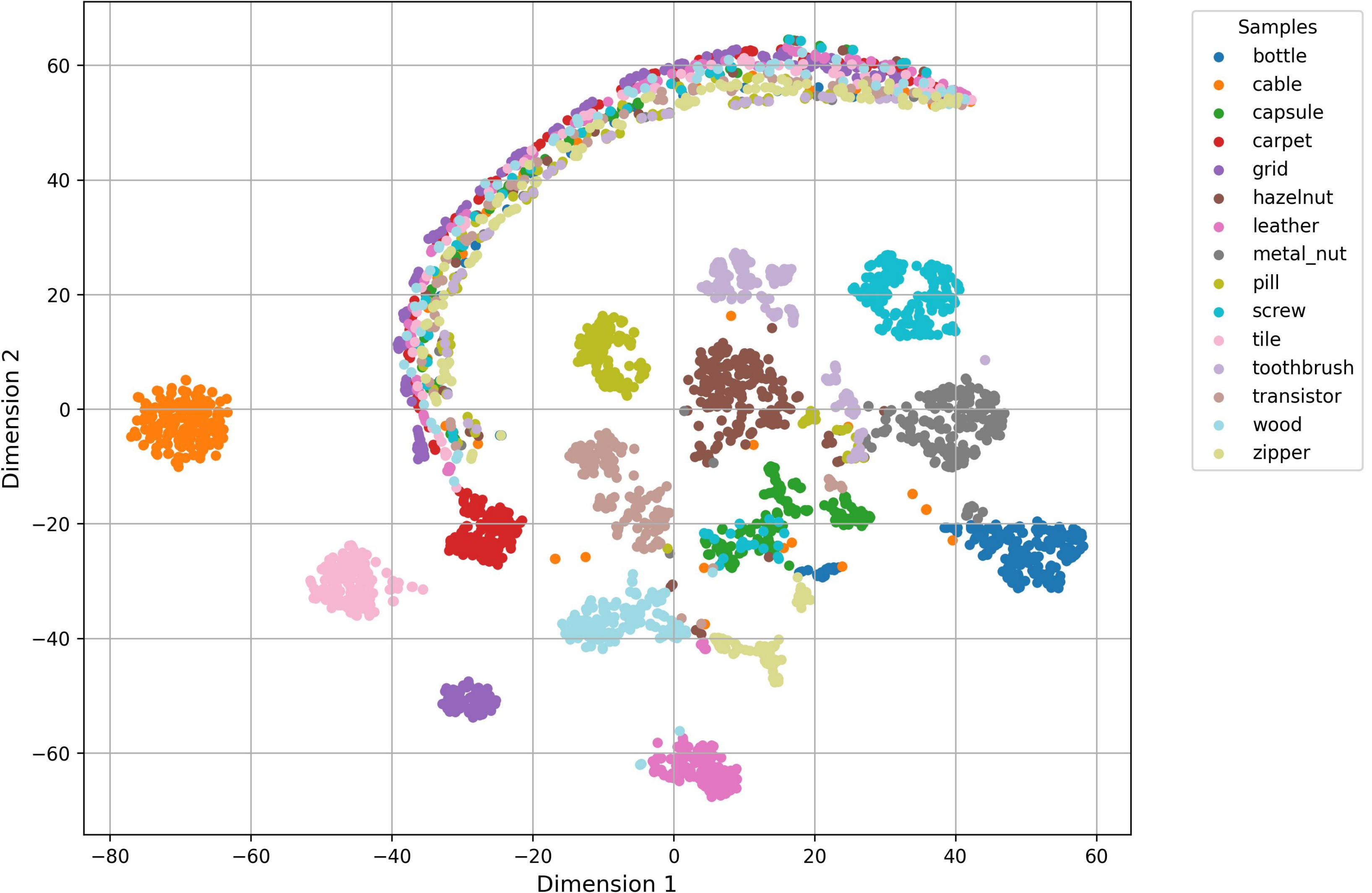} 
	\caption{The T-SNE Visualization results of MTRMB. The dimensionality reduction visualization results of 15 classes show that the proposed method can generate a compact task representation memory bank for UCAD. It should be noted that the arc\-shaped dimensionality reduction feature sets are all generated by the background area. This is because the background areas of each category in the MVTec AD dataset are very similar.}
	\label{tsne} 
\end{figure}
\textbf{Methods:}Based on the anomaly methods discussed in our related work section and previous benchmark \cite{xie2024iad}, we selected the most representative methods from each paradigm to establish the benchmark. These methods include  CutPaste \cite{li2021cutpaste}, CSFlow \cite{rudolph2022fully},
Fastflow \cite{yu2021fastflow},
FAVAE \cite{dehaene2020anomaly},
SimpleNet \cite{liu2023simplenet},
DRAEM \cite{zavrtanik2021draem},
PaDiM \cite{defard2021padim},
SPADE \cite{cohen2020sub},
STPM \cite{salehi2021multiresolution},
PatchCore \cite{roth2022towards},
CFA \cite{lee2022cfa},
DNE \cite{li2022towards}, 
RD4AD \cite{deng2022anomaly}, 
UniAD \cite{you2022unified} and UCAD \cite{liu2024unsupervised}.

\begin{figure}[t]
	\includegraphics[width=0.8\columnwidth, trim=0 0 0 0, clip]{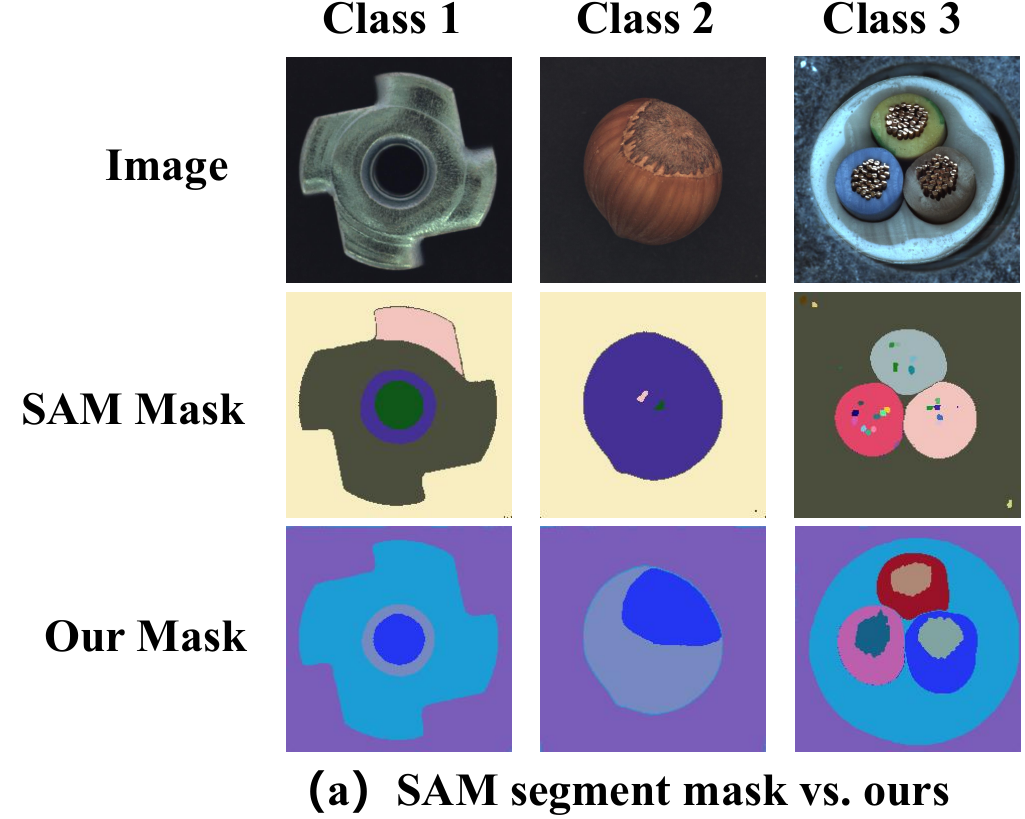} 
	\caption{The first row shows the test images, the second row shows the structure regions generated by SAM, and the third row shows the structure regions generated by the effective interaction between only Grounding DINO and SAM. The segmented region visualization shows that our method can generate more accurate structure regions to guide the model to learn the key parts of the object.}
	\label{gd} 
\end{figure}

\textbf{Metrics:}Following the common practice, we utilize Area Under the Receiver Operating Characteristics (AUROC/AUC) to assess the model’s ability in anomaly classification. For pixel-level anomaly segmentation capability, we employ Area Under Precision-Recall (AUPR/AP) \cite{bergmann2020uninformed} for model evaluation. In addition, we use Forgetting Measure(FM) \cite{chaudhry2018riemannian} to evaluate models’ ability to prevent catastrophic forgetting.

\begin{equation}
    \resizebox{.91\linewidth}{!}{$
            \displaystyle
            avg FM = \frac{1}{(k-1)k} \sum_{j=1}^{k-1} \mathop{max}\limits_{l\in\{1,...,k-1\}} T_{l,j}-T_{k,j}
        $}
\end{equation}

where \textbf{$T$} represents tasks, $k$ stands for the current training task ID, and $j$ refers to the task ID being evaluated. And $avg FM$ represents the average forgetting measure of the model after completing $k$ tasks. During the inference, we evaluate the model after training on all tasks.



\begin{table*}[t]
\centering
\resizebox{\textwidth}{!}{
\begin{tabular}{lccccccccccccccccc}
\hline
Methods  & Bottle & Cable & Capsule & Carpet& Grid& Hazelnut & Leather & Metal$\_$Nut& Pill& Screw & Tile& Toothbrush & Transistor& Wood & Zipper & Average& Avg FM\\
\hline
CFA & 0.309  & 0.489 & 0.275 & 0.834 & 0.571 & 0.903 & 0.935 & 0.464 & 0.528 & 0.528 & 0.763 & 0.519 & 0.320 & 0.923 & 0.984 & 0.623 & 0.361\\
CSFlow & 0.129  & 0.420 & 0.363 & 0.978 & 0.602 & 0.269 & 0.906 & 0.220 & 0.263 & 0.434 & 0.697 & 0.569 & 0.432 & 0.802 & 0.997 & 0.539 & 0.426\\
CutPaste  & 0.111& 0.422 & 0.373 & 0.198& 0.214& 0.578 & 0.007 & 0.517& 0.371& 0.356 & 0.112& 0.158 & 0.340& 0.150 & 0.775 & 0.312& 0.510\\
DRAEM  & 0.793& 0.411 & 0.517& 0.537& 0.799& 0.524 & 0.480 & 0.422& 0.452& 1.000 & 0.548& 0.625 & 0.307& 0.517 & 0.996 & 0.595& 0.371\\
FastFlow  & 0.454& 0.512 & 0.517& 0.489& 0.482& 0.522 & 0.487 & 0.476& 0.575& 0.402 & 0.489& 0.267 & 0.526& 0.616 & 0.867 & 0.512& 0.279\\
FAVAE  & 0.666& 0.396 & 0.357&0.610& 0.644& 0.884 & 0.406 & 0.416& 0.531& 0.624 & 0.563& 0.503 & 0.331& 0.728 &0.544 & 0.547& 0.102\\
PaDiM & 0.458 & 0.544  & 0.418 & 0.454 & 0.704 & 0.635 & 0.418 & 0.446 & 0.449 & 0.578 & 0.581 & 0.678 & 0.407 & 0.549 & 0.855 & 0.545 & 0.368\\
Patchcore  & 0.163 & 0.518 & 0.350 & 0.968 & 0.700 & 0.839 & 0.625 & 0.259 & 0.459 & 0.484 & 0.776 & 0.586 & 0.341 & 0.970 & 0.991 & 0.602 & 0.383\\
RD4AD  & 0.401 & 0.538 & 0.475 & 0.583 & 0.558 & 0.909 & 0.596 & 0.623 & 0.479 & 0.596 & 0.715 & 0.397 & 0.385 & 0.700 & 0.987 & 0.596 & 0.285\\
SPADE  & 0.302 & 0.444 & 0.525 & 0.529 & 0.460 & 0.410 & 0.577 & 0.592 & 0.484 & 0.514 & 0.881 & 0.386 & 0.622 & 0.897 & 0.949 & 0.571& 0.393\\
STPM & 0.329 & 0.539  & 0.610 & 0.462 & 0.569 & 0.540 & 0.740& 0.456 & 0.523 & 0.753 & 0.736 & 0.375 & 0.450 & 0.779 & 0.783 & 0.576 & 0.325\\
SimpleNet & 0.938 & 0.560  & 0.519 & 0.736 & 0.592 & 0.859 & 0.749 & 0.710 & 0.701 & 0.599 & 0.654 & 0.422 & 0.669 & 0.908 & 0.996 & 0.708 & 0.211\\
UniAD  & 0.801  & 0.660 & 0.823 & 0.754 & 0.713 & 0.904 & 0.715 & 0.791 & 0.869 & 0.731 & 0.687 & 0.776 & 0.490 & 0.903 & 0.997 & 0.774 & 0.229\\
\hline
Patchcore$^*$ & 0.533  & 0.505 & 0.351  & 0.865&0.723& 0.959 & 0.854 & 0.456& 0.511& 0.626 & 0.748& 0.600 & 0.427& 0.900 & 0.974 & 0.669& 0.318\\
UniAD$^*$  & 0.997  & 0.701 & 0.765 & \textbf{0.998} & 0.896 & 0.936 & \textbf{1.000} & 0.964 & 0.895 & 0.554 & 0.989 & 0.928 & \textbf{0.966} & 0.982 & \textbf{0.987} & 0.904 & 0.076\\
DNE & 0.990 & 0.619 & 0.609 & 0.984 & \textbf{0.998} & 0.924 & \textbf{1.000} & 0.989 & 0.671 & 0.588 & 0.980 & 0.933 & 0.877 & 0.930 & 0.958 & 0.870 & 0.116\\
UCAD &  0.995  & \textbf{0.731}  &  0.866  & 0.965 & 0.944& \textbf{0.994}& 0.996 & 0.988  & 0.890 &0.739 &  \textbf{0.998}& 0.978 & 0.874 & \textbf{0.995} & 0.938 & 0.926 & 0.056 \\
\hline
Ours & \textbf{1.000} & 0.652 & \textbf{0.926} & 0.970 & 0.976 & 0.990 & \textbf{1.000} & \textbf{0.995} & \textbf{0.892} & \textbf{0.852} & 0.997 & \textbf{1.000} & 0.926 & 0.991 & 0.949 & \textbf{0.941} & \textbf{0.016}\\
\bottomrule  
\end{tabular}
}
\caption{Image-level $AUROC\uparrow$ and corrsponding $FM\downarrow$ on MVTec AD dataset after training on the last subdataset. The best results are highlighted in bold.}
\label{mvtec_image_auc}
\end{table*}

\begin{table*}[t]
\centering
\resizebox{\textwidth}{!}{
\begin{tabular}{lccccccccccccccccc}
\hline
Methods  & Bottle & Cable & Capsule & Carpet& Grid& Hazelnut & Leather & Metal$\_$Nut& Pill& Screw & Tile& Toothbrush & Transistor& Wood & Zipper & Average& Avg FM\\
\hline
CFA & 0.068  & 0.056 & 0.050 & 0.271 & 0.004 & 0.341 & 0.393 & 0.255 & 0.080 & 0.015 & 0.155 & 0.053 & 0.056 & 0.281 & 0.573 & 0.177 & 0.083\\
DRAEM & 0.117  & 0.019 & 0.044 & 0.018 & 0.005 & 0.036 & 0.013 & 0.142 & 0.104 & 0.002 & 0.130 & 0.039 & 0.040 & 0.033 & 0.734 & 0.098 & 0.116\\
FastFlow & 0.044  & 0.021 & 0.013 & 0.013 & 0.005 & 0.028 & 0.007 & 0.090 & 0.029 & 0.003 & 0.060 & 0.015 & 0.036 & 0.037 & 0.264 & 0.044 & 0.214\\
FAVAE & 0.086  & 0.048 & 0.039 & 0.015 & 0.004 & 0.389 & 0.112 & 0.174 & 0.070 & 0.017 & 0.064 & 0.043 & 0.046 & 0.093 & 0.039 & 0.083 & 0.083\\
PaDim  & 0.072 & 0.037 & 0.030 & 0.023 & 0.006 & 0.183 & 0.039 & 0.155 & 0.044 & 0.014 & 0.065 & 0.044 & 0.049 & 0.080 & 0.452 & 0.086 & 0.366\\
Patchcore  & 0.048 & 0.029 & 0.035 & 0.552 & 0.003 & 0.338 & 0.279 & 0.248 & 0.051 & 0.008 & 0.249 & 0.034 & 0.079 & 0.304 & 0.595 & 0.190 & 0.371\\
RD4AD  & 0.055 & 0.040 & 0.064 & 0.212 & 0.005 & 0.384 & 0.116 & 0.247 & 0.061 & 0.015 & 0.193 & 0.034 & 0.059 & 0.097 & 0.562 & 0.143 & 0.425\\
SPADE & 0.122 & 0.052 & 0.044 & 0.117 & 0.004 & 0.512 & 0.264 & 0.181 & 0.060 & 0.020 & 0.096 & 0.043 & 0.050 & 0.172 & 0.531 & 0.151 & 0.319\\
STPM & 0.074 & 0.019 & 0.073 & 0.054 & 0.005 & 0.037 & 0.108 & 0.354 & 0.111 & 0.001 & 0.397 & 0.046 & 0.046 & 0.119 & 0.203 & 0.110 & 0.352\\
SimpleNet & 0.108 & 0.045 & 0.029 & 0.018 & 0.004 & 0.029 & 0.006 & 0.227 & 0.077 & 0.004 & 0.082 & 0.046 & 0.049 & 0.037 & 0.139 & 0.060 & 0.069\\
UniAD  & 0.734 & 0.232 & 0.313 & 0.517 & 0.204 & 0.378 & 0.360 & 0.587 & 0.346 & 0.035 & 0.428 & 0.398 & 0.542& 0.378 & 0.443 & 0.393 & 0.086\\
\hline
UniAD$^*$  & 0.054 & 0.031 & 0.022 & 0.047 & 0.007 & 0.189 & 0.053 & 0.110 & 0.034 & 0.008 & 0.107& 0.040 & 0.045& 0.103 & 0.444 & 0.086 & 0.419\\
Patchcore$^*$ & 0.087 & 0.043 & 0.042 & 0.407& 0.003 & 0.443 & \textbf{0.352} & 0.189& 0.058& 0.017 & 0.124& 0.028 & 0.053& 0.270 & \textbf{0.604} & 0.181& 0.343\\
UCAD &  0.751  & \textbf{0.271}  &  0.339  & 0.622 &0.185 & 0.506  & 0.333  & 0.765 & \textbf{0.634}& 0.214  & \textbf{0.549}& 0.288  & 0.398 & 0.535 & 0.388 & 0.451 & 0.023 \\
\hline
Ours & \textbf{0.753} & 0.244 & \textbf{0.344} & \textbf{0.637} & \textbf{0.190} & \textbf{0.519} & 0.337 & \textbf{0.811} & 0.631 & \textbf{0.232} & 0.546 & \textbf{0.301} & \textbf{0.483}& \textbf{0.573} & 0.420 & \textbf{0.468}& \textbf{0.017}\\
\hline
\end{tabular}
}
\caption{Pixel-level $AUPR\uparrow$ and corrsponding $FM\downarrow$ on MVTec AD dataset after training on the last subdataset.}
\label{mvtec_pixel_pr}
\end{table*}

\begin{figure*}[t]
	\centering 
	\includegraphics[scale=0.28]{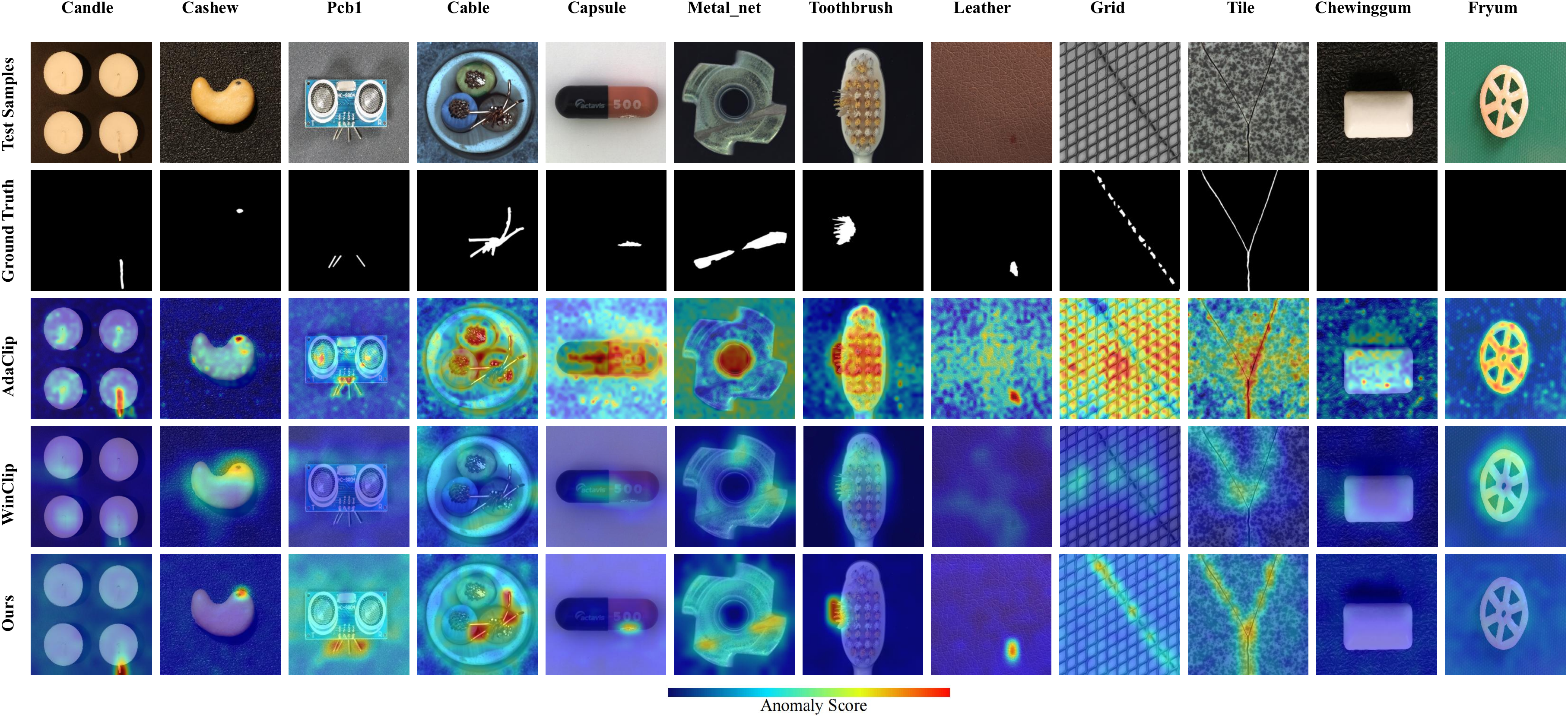} 
	\caption{Visualization examples of continual anomaly detection. The first row displays the original anomaly images, the second row shows the ground truth annotations, and the third to fifth rows depict the heatmaps of our method and other methods.}
	\label{vis} 
\end{figure*}

\begin{table*}[t]
\centering
\resizebox{\textwidth}{!}{
\begin{tabular}{lccccccccccccccccc}
\hline
Methods  & Candle & Capsules & Cashew & Chewinggum & Fryum & Macaroni1 & Macaroni2 & Pcb1 & Pcb2 & Pcb3 & Pcb4 & Pipe$\_$fryum & Average & AvgFM\\
\hline
CFA & 0.512   & 0.672 & 0.873 & 0.753 & 0.304 & 0.557 & 0.422 & 0.698 & 0.472& 0.449 & 0.407 & 0.998 & 0.593& 0.327\\
RD4AD     & 0.380 & 0.385 & 0.737 & 0.539 & 0.533 & 0.607 & 0.487 & 0.437& 0.672 & 0.343 & 0.187 & 0.999 & 0.525 & 0.423\\
Patchcore  & 0.401 & 0.605 & 0.624 & 0.907 & 0.334 & 0.538 & 0.437 & 0.527& 0.597 & 0.507 & 0.588 & 0.998 & 0.589 & 0.361\\
SimpleNet  & 0.504 & 0.474  & 0.794 & 0.721 & 0.684 & 0.567 & 0.447 & 0.598 & 0.629 & 0.538 & 0.493 & 0.945 & 0.616 & 0.283\\
UniAD  & 0.573  & 0.599 & 0.661 & 0.758 & 0.504 & 0.559 & 0.644 & 0.749 & 0.523 & 0.547 & 0.562 & 0.989 & 0.639 & 0.297\\
\hline
UniAD$^*$  & 0.884  & 0.669 & 0.938 & 0.970 & 0.812 & 0.753 & 0.570 & 0.872 & 0.766 & 0.708 & 0.967 & 0.990 & 0.825 & 0.125\\
DNE & 0.486 & 0.413 & 0.735 & 0.585 & 0.691 & 0.584 & 0.546 & 0.633 & 0.693& 0.642 & 0.562 & 0.747 & 0.610 & 0.179\\
Patchcore$^*$  & 0.647 & 0.579 & 0.669 & 0.735 & 0.431 & 0.631 & 0.624 & 0.617& 0.534 & 0.479 & 0.645 &  \textbf{0.999} & 0.633 & 0.349\\
UCAD &  0.778  & 0.877  &   \textbf{0.960}  & 0.958 &  \textbf{0.945}&  \textbf{0.823}  &  0.667 & \textbf{0.905}  &0.871 & 0.813  & 0.901& 0.988  & 0.874 &0.039 \\
\hline
Ours &  \textbf{0.946} &  \textbf{0.937} & 0.935 &  \textbf{0.989} & 0.874 & 0.783 &  \textbf{0.794} & 0.765&  \textbf{0.914}&  \textbf{0.938} &  \textbf{0.997}& 0.925 &  \textbf{0.900}&  \textbf{0.010} \\
\hline
\end{tabular}
}
\caption{Image-level $AUROC\uparrow$ and corrsponding $FM\downarrow$ on VisA dataset after training on the last subdataset.}
\label{visa_image_auc}
\end{table*}


\subsection{Continual Anomaly Detection Benchmark}
We conducted comprehensive evaluations of the aforementioned 15 methods on the MVTec AD and VisA datasets. Among them, UCAD stands as the SOTA method in unsupervised continual AD. Meanwhile, DNE and UniAD are two representative AD methods for continual and unified methods, respectively. Intuitively, these two methods appear to be better suited for the open learning scenario. Due to the famous replay in continual learning methods, we also conducted replay-based experiments on PatchCore and UniAD. Comprehensive experiments verify the superiority of our approach in persistent anomaly detection.

\subsubsection{Quantitative Analysis}
As shown in Table \ref{mvtec_image_auc}-\ref{visa_image_auc}, our method outperforms other anomaly detection algorithms in both continuous detection and segmentation on the MVtec AD and VisA datasets. The detection and segmentation effects of common anomaly detection methods are greatly reduced in the continuous detection scenario. Since there are few AD studies with continuous anomalies, we add a playback mechanism to Patchcore and UniAD to facilitate further comparison. However, $Patchcore^*$ with continuous learning ability is still far lower than the proposed method in Image AUROC and Pixel AUROC by 0.27, 0.287 (MVTec AD), and $UniAD^*$ is 0.041 and 0.382 lower than our method, respectively. In addition, we also fully compared with the most advanced continuous anomaly detection methods DNE and UCAD. Our method is superior to DNE and UCAD in continuous detection ability while maintaining the lowest forgetting rate. On the MVTec AD dataset, compared with DNE and UCAD, our method has an Image AUROC that is 0.071 and 0.015 higher, respectively. On the VisA dataset, compared with DNE and UCAD, our method has an Image AUROC that is 0.29 and 0.026 higher, respectively. Comprehensive experiments show that our method can effectively resist catastrophic forgetting, and the proposed method has superior continuous anomaly detection capabilities. We attribute this to the task representation memory bank we designed, which obtains comprehensive task representations for continuous learning and unsupervised detection by strengthening the interaction of multimodal information.

\subsubsection{Qualitative Analysis}
As shown in Fig.\ref{vis}, our method shows two obvious advantages over ADClip and UniAD. First, it demonstrates more accurate anomaly localization. Second, it minimizes false positives in normal image classification. As shown in Fig.\ref{gd}, our method is able to obtain better structural regions for contrastive learning, which greatly improves the model's learning of contextual features. To verify that the proposed method can generate an effective task representation memory bank, we performed T-SNE visualization of the memory bank after training 15 categories in MVTec AD, as shown in Figure\ref{tsne}. The results show that the proposed method can generate a task representation memory bank that is compact and not easy to forget knowledge (after incremental learning of 15 categories, the features of different categories are not confused together) for continuous learning and detection.


\subsection{Ablation Study}

\begin{table}[ht]
\resizebox{\columnwidth}{!}{
\begin{tabular}{cccccc}
\toprule
RSCL  & KPMK & GD & Image\_AUC & Pixel\_AP \\
\midrule
\ding{55} & \ding{55} & \ding{55} & 0.894 & 0.426 \\
\checkmark & \ding{55} & \ding{55} & 0.924 & 0.447 \\
\checkmark & \checkmark & \ding{55} & 0.935 & 0.458 \\
\checkmark & \checkmark & \checkmark & \textbf{0.941} & \textbf{0.468} \\
\bottomrule
\end{tabular}
}
\caption{Ablation experiments of key modules (RSCL, KPMK, GD), GD is the abbreviation of Grounding DINO. We calculated on the average detection and segmentation metrics of MVTec AD and VisA.}

\label{modules}
\end{table}

\subsubsection{Module Effectivity}
As shown in Table \ref{modules}, we analyze the impact of three parts: KPMK, RSCL, and Grounding DINO. We observe that the performance of the model is significantly improved with the implementation of these modules. Regarding RSCL, we find that RSCL can effectively improve the continuous detection ability. The addition of this part can improve $Image\_AUC$ and $Pixel\_AP$ by 0.03 and 0.021 respectively. Without KPMK, our model uses a single-modal knowledge base and cannot be fully represented each time a new task is introduced. This approach limits the model's ability to continuously learn without supervision. KPMK improves $Image\_AUC$ and $Pixel\_AP$ by 0.011 and 0.011 respectively by strengthening the interaction of different modal information. By adding Grounding DINO, the model can be guided to better learn contextual features, which improves $Image\_AUC$ and $Pixel\_AP$ by 0.006 and 0.01 respectively. The experimental results show that the key components proposed in this paper can effectively improve the model's continuous learning ability in AD.


\begin{table}[h]
    \resizebox{\columnwidth}{!}{
    \begin{tabular}{cccccc}
        \toprule
        Pt\_Len & Img\_AUC & Pix\_AUC & Img\_AP & Pix\_AP & Pix\_PRO \\
        \midrule
        1  & 0.936 & 0.971 & 0.962 & 0.472 & 0.900 \\
        5  & \textbf{0.941} & \textbf{0.976} & \textbf{0.965} & \textbf{0.488} & \textbf{0.908} \\
        10 & 0.931 & 0.975 & 0.958 & 0.457 & 0.903 \\
        15 & 0.925 & 0.974 & 0.955 & 0.470 & 0.903 \\
        \bottomrule
    \end{tabular}
    }
    \caption{Ablation experiments on prompt size hyperparameters. We calculated on the average detection and segmentation metrics of MVTec AD and VisA.}
    \label{hp}
\end{table}

\subsubsection{Impact of Prompt Hyperparameterss}
To further study the influence of hyperparameters in learnable prompts, we designed an ablation experiment as shown in Table \ref{hp}. By changing the length of the prompt, we observe the trend of changes in the five detection indicators to select the optimal hyperparameters. The prompt length is set to 1, 5, 10, and 15 respectively. From the results, the prompt length of 5 has the best detection and segmentation accuracy, and its excess over the suboptimal situation on $Image\_AUC$ and $Pixel\_AP$ is 0.005 and 0.016 respectively. The size of the prompt directly affects the effect of continuous learning, and choosing the appropriate prompt hyperparameters is critical.


\section*{Conclusion}
In this paper, we study the problem of applying incremental learning to unsupervised anomaly detection for practical applications in industrial manufacturing. To facilitate this research, we establish a comprehensive benchmark for unsupervised persistent anomaly detection and segmentation. In addition, our proposed MTRMB for the UCAD task is effective against catastrophic forgetting, which is the first study to apply multimodal prompt for incremental learning to unsupervised anomaly detection. The novelty of MTRMB relies on the KPMK mechanism and RSML, which utilizes the interaction of multimodal information to construct a compact task representation memory bank, which significantly improves the performance of persistent anomaly detection. Comprehensive experiments highlight the effectiveness and robustness of our framework under different hyperparameters. We also find that the persistent detection effect can be further improved by designing the prompt, which is an important direction for our subsequent optimization.

\appendix



\bibliographystyle{named}

\begin{thebibliography}{10}
\providecommand{\url}[1]{#1}
\csname url@samestyle\endcsname
\providecommand{\newblock}{\relax}
\providecommand{\bibinfo}[2]{#2}
\providecommand{\BIBentrySTDinterwordspacing}{\spaceskip=0pt\relax}
\providecommand{\BIBentryALTinterwordstretchfactor}{4}
\providecommand{\BIBentryALTinterwordspacing}{\spaceskip=\fontdimen2\font plus
\BIBentryALTinterwordstretchfactor\fontdimen3\font minus \fontdimen4\font\relax}
\providecommand{\BIBforeignlanguage}[2]{{%
\expandafter\ifx\csname l@#1\endcsname\relax
\typeout{** WARNING: IEEEtran.bst: No hyphenation pattern has been}%
\typeout{** loaded for the language `#1'. Using the pattern for}%
\typeout{** the default language instead.}%
\else
\language=\csname l@#1\endcsname
\fi
#2}}
\providecommand{\BIBdecl}{\relax}
\BIBdecl

\bibitem{bergmann2019mvtec}
P.~Bergmann, M.~Fauser, D.~Sattlegger, and C.~Steger, ``Mvtec ad--a comprehensive real-world dataset for unsupervised anomaly detection,'' in \emph{Proceedings of the IEEE/CVF conference on computer vision and pattern recognition}, 2019, pp. 9592--9600.

\bibitem{zou2022spot}
Y.~Zou, J.~Jeong, L.~Pemula, D.~Zhang, and O.~Dabeer, ``Spot-the-difference self-supervised pre-training for anomaly detection and segmentation,'' \emph{arXiv preprint arXiv:2207.14315}, 2022.

\bibitem{liu2024deep}
J.~Liu, G.~Xie, J.~Wang, S.~Li, C.~Wang, F.~Zheng, and Y.~Jin, ``Deep industrial image anomaly detection: A survey,'' \emph{Machine Intelligence Research}, vol.~21, no.~1, pp. 104--135, 2024.

\bibitem{xie2024iad}
G.~Xie, J.~Wang, J.~Liu, J.~Lyu, Y.~Liu, C.~Wang, F.~Zheng, and Y.~Jin, ``Im-iad: Industrial image anomaly detection benchmark in manufacturing,'' \emph{IEEE Transactions on Cybernetics}, 2024.

\bibitem{lee2022cfa}
S.~Lee, S.~Lee, and B.~C. Song, ``Cfa: Coupled-hypersphere-based feature adaptation for target-oriented anomaly localization,'' \emph{IEEE Access}, vol.~10, pp. 78\,446--78\,454, 2022.

\bibitem{gudovskiy2022cflow}
D.~Gudovskiy, S.~Ishizaka, and K.~Kozuka, ``Cflow-ad: Real-time unsupervised anomaly detection with localization via conditional normalizing flows,'' in \emph{Proceedings of the IEEE/CVF winter conference on applications of computer vision}, 2022, pp. 98--107.

\bibitem{rudolph2022fully}
M.~Rudolph, T.~Wehrbein, B.~Rosenhahn, and B.~Wandt, ``Fully convolutional cross-scale-flows for image-based defect detection,'' in \emph{Proceedings of the IEEE/CVF Winter Conference on Applications of Computer Vision}, 2022, pp. 1088--1097.

\bibitem{li2021cutpaste}
C.-L. Li, K.~Sohn, J.~Yoon, and T.~Pfister, ``Cutpaste: Self-supervised learning for anomaly detection and localization,'' in \emph{Proceedings of the IEEE/CVF conference on computer vision and pattern recognition}, 2021, pp. 9664--9674.

\bibitem{zhang2023iddm}
F.~Zhang and Z.~Chen, ``Iddm: An incremental dual-network detection model for in-situ inspection of large-scale complex product,'' \emph{Journal of Industrial Information Integration}, vol.~33, p. 100463, 2023.

\bibitem{li2022towards}
W.~Li, J.~Zhan, J.~Wang, B.~Xia, B.-B. Gao, J.~Liu, C.~Wang, and F.~Zheng, ``Towards continual adaptation in industrial anomaly detection,'' in \emph{Proceedings of the 30th ACM International Conference on Multimedia}, 2022, pp. 2871--2880.

\bibitem{li2023cross}
W.~Li, B.-B. Gao, B.~Xia, J.~Wang, J.~Liu, Y.~Liu, C.~Wang, and F.~Zheng, ``Cross-modal alternating learning with task-aware representations for continual learning,'' \emph{IEEE Transactions on Multimedia}, 2023.

\bibitem{zhang2024realnet}
X.~Zhang, M.~Xu, and X.~Zhou, ``Realnet: A feature selection network with realistic synthetic anomaly for anomaly detection,'' in \emph{Proceedings of the IEEE/CVF Conference on Computer Vision and Pattern Recognition}, 2024, pp. 16\,699--16\,708.

\bibitem{roth2022towards}
K.~Roth, L.~Pemula, J.~Zepeda, B.~Sch{\"o}lkopf, T.~Brox, and P.~Gehler, ``Towards total recall in industrial anomaly detection,'' in \emph{Proceedings of the IEEE/CVF conference on computer vision and pattern recognition}, 2022, pp. 14\,318--14\,328.

\bibitem{bergmann2020uninformed}
P.~Bergmann, M.~Fauser, D.~Sattlegger, and C.~Steger, ``Uninformed students: Student-teacher anomaly detection with discriminative latent embeddings,'' in \emph{Proceedings of the IEEE/CVF conference on computer vision and pattern recognition}, 2020, pp. 4183--4192.

\bibitem{salehi2021multiresolution}
M.~Salehi, N.~Sadjadi, S.~Baselizadeh, M.~H. Rohban, and H.~R. Rabiee, ``Multiresolution knowledge distillation for anomaly detection,'' in \emph{Proceedings of the IEEE/CVF conference on computer vision and pattern recognition}, 2021, pp. 14\,902--14\,912.

\bibitem{deng2022anomaly}
H.~Deng and X.~Li, ``Anomaly detection via reverse distillation from one-class embedding,'' in \emph{Proceedings of the IEEE/CVF conference on computer vision and pattern recognition}, 2022, pp. 9737--9746.

\bibitem{tien2023revisiting}
T.~D. Tien, A.~T. Nguyen, N.~H. Tran, T.~D. Huy, S.~Duong, C.~D.~T. Nguyen, and S.~Q. Truong, ``Revisiting reverse distillation for anomaly detection,'' in \emph{Proceedings of the IEEE/CVF conference on computer vision and pattern recognition}, 2023, pp. 24\,511--24\,520.

\bibitem{batzner2024efficientad}
K.~Batzner, L.~Heckler, and R.~K{\"o}nig, ``Efficientad: Accurate visual anomaly detection at millisecond-level latencies,'' in \emph{Proceedings of the IEEE/CVF Winter Conference on Applications of Computer Vision}, 2024, pp. 128--138.

\bibitem{liu2025grounding}
S.~Liu, Z.~Zeng, T.~Ren, F.~Li, H.~Zhang, J.~Yang, Q.~Jiang, C.~Li, J.~Yang, H.~Su \emph{et~al.}, ``Grounding dino: Marrying dino with grounded pre-training for open-set object detection,'' in \emph{European Conference on Computer Vision}.\hskip 1em plus 0.5em minus 0.4em\relax Springer, 2025, pp. 38--55.

\bibitem{kirillov2023segment}
A.~Kirillov, E.~Mintun, N.~Ravi, H.~Mao, C.~Rolland, L.~Gustafson, T.~Xiao, S.~Whitehead, A.~C. Berg, W.-Y. Lo \emph{et~al.}, ``Segment anything,'' in \emph{Proceedings of the IEEE/CVF International Conference on Computer Vision}, 2023, pp. 4015--4026.

\bibitem{jeong2023winclip}
J.~Jeong, Y.~Zou, T.~Kim, D.~Zhang, A.~Ravichandran, and O.~Dabeer, ``Winclip: Zero-/few-shot anomaly classification and segmentation,'' in \emph{Proceedings of the IEEE/CVF Conference on Computer Vision and Pattern Recognition}, 2023, pp. 19\,606--19\,616.

\bibitem{cao2025adaclip}
Y.~Cao, J.~Zhang, L.~Frittoli, Y.~Cheng, W.~Shen, and G.~Boracchi, ``Adaclip: Adapting clip with hybrid learnable prompts for zero-shot anomaly detection,'' in \emph{European Conference on Computer Vision}.\hskip 1em plus 0.5em minus 0.4em\relax Springer, 2025, pp. 55--72.

\bibitem{chen2023april}
X.~Chen, Y.~Han, and J.~Zhang, ``April-gan: A zero-/few-shot anomaly classification and segmentation method for cvpr 2023 vand workshop challenge tracks 1\&2: 1st place on zero-shot ad and 4th place on few-shot ad,'' \emph{arXiv preprint arXiv:2305.17382}, 2023.

\bibitem{cao2023winning}
Y.~Cao, X.~Xu, C.~Sun, Y.~Cheng, L.~Gao, and W.~Shen, ``Winning solution for the cvpr2023 visual anomaly and novelty detection challenge: Multimodal prompting for data-centric anomaly detection,'' \emph{arXiv preprint arXiv:2306.09067}, 2023.

\bibitem{oquab2023dinov2}
M.~Oquab, T.~Darcet, T.~Moutakanni, H.~Vo, M.~Szafraniec, V.~Khalidov, P.~Fernandez, D.~Haziza, F.~Massa, A.~El-Nouby \emph{et~al.}, ``Dinov2: Learning robust visual features without supervision,'' \emph{arXiv preprint arXiv:2304.07193}, 2023.

\bibitem{yi2020patch}
J.~Yi and S.~Yoon, ``Patch svdd: Patch-level svdd for anomaly detection and segmentation,'' in \emph{Proceedings of the Asian conference on computer vision}, 2020.

\bibitem{li2024promptad}
X.~Li, Z.~Zhang, X.~Tan, C.~Chen, Y.~Qu, Y.~Xie, and L.~Ma, ``Promptad: Learning prompts with only normal samples for few-shot anomaly detection,'' in \emph{Proceedings of the IEEE/CVF Conference on Computer Vision and Pattern Recognition}, 2024, pp. 16\,838--16\,848.

\bibitem{massoli2021mocca}
F.~V. Massoli, F.~Falchi, A.~Kantarci, {\c{S}}.~Akti, H.~K. Ekenel, and G.~Amato, ``Mocca: Multilayer one-class classification for anomaly detection,'' \emph{IEEE transactions on neural networks and learning systems}, vol.~33, no.~6, pp. 2313--2323, 2021.

\bibitem{liu2023simplenet}
Z.~Liu, Y.~Zhou, Y.~Xu, and Z.~Wang, ``Simplenet: A simple network for image anomaly detection and localization,'' in \emph{Proceedings of the IEEE/CVF Conference on Computer Vision and Pattern Recognition}, 2023, pp. 20\,402--20\,411.

\bibitem{cao2023anomaly}
T.~Cao, J.~Zhu, and G.~Pang, ``Anomaly detection under distribution shift,'' in \emph{Proceedings of the IEEE/CVF International Conference on Computer Vision}, 2023, pp. 6511--6523.

\bibitem{zhou2024msflow}
Y.~Zhou, X.~Xu, J.~Song, F.~Shen, and H.~T. Shen, ``Msflow: Multiscale flow-based framework for unsupervised anomaly detection,'' \emph{IEEE Transactions on Neural Networks and Learning Systems}, 2024.

\bibitem{xie2023pushing}
G.~Xie, J.~Wang, J.~Liu, F.~Zheng, and Y.~Jin, ``Pushing the limits of fewshot anomaly detection in industry vision: Graphcore,'' \emph{arXiv preprint arXiv:2301.12082}, 2023.

\bibitem{you2022unified}
Z.~You, L.~Cui, Y.~Shen, K.~Yang, X.~Lu, Y.~Zheng, and X.~Le, ``A unified model for multi-class anomaly detection,'' \emph{Advances in Neural Information Processing Systems}, vol.~35, pp. 4571--4584, 2022.

\bibitem{zhao2023omnial}
Y.~Zhao, ``Omnial: A unified cnn framework for unsupervised anomaly localization,'' in \emph{Proceedings of the IEEE/CVF Conference on Computer Vision and Pattern Recognition}, 2023, pp. 3924--3933.

\bibitem{ho2024long}
C.-H. Ho, K.-C. Peng, and N.~Vasconcelos, ``Long-tailed anomaly detection with learnable class names,'' in \emph{Proceedings of the IEEE/CVF Conference on Computer Vision and Pattern Recognition}, 2024, pp. 12\,435--12\,446.

\bibitem{schluter2022natural}
H.~M. Schl{\"u}ter, J.~Tan, B.~Hou, and B.~Kainz, ``Natural synthetic anomalies for self-supervised anomaly detection and localization,'' in \emph{European Conference on Computer Vision}.\hskip 1em plus 0.5em minus 0.4em\relax Springer, 2022, pp. 474--489.

\bibitem{d2023multimodal}
M.~D'Alessandro, A.~Alonso, E.~Calabr{\'e}s, and M.~Galar, ``Multimodal parameter-efficient few-shot class incremental learning,'' in \emph{Proceedings of the IEEE/CVF International Conference on Computer Vision}, 2023, pp. 3393--3403.

\bibitem{zavrtanik2021draem}
V.~Zavrtanik, M.~Kristan, and D.~Sko{\v{c}}aj, ``Draem-a discriminatively trained reconstruction embedding for surface anomaly detection,'' in \emph{Proceedings of the IEEE/CVF international conference on computer vision}, 2021, pp. 8330--8339.

\bibitem{peng2024industrial}
J.~Peng, H.~Shao, Y.~Xiao, B.~Cai, and B.~Liu, ``Industrial surface defect detection and localization using multi-scale information focusing and enhancement ganomaly,'' \emph{Expert Systems with Applications}, vol. 238, p. 122361, 2024.

\bibitem{yao2024prior}
H.~Yao, Y.~Cao, W.~Luo, W.~Zhang, W.~Yu, and W.~Shen, ``Prior normality prompt transformer for multi-class industrial image anomaly detection,'' \emph{arXiv preprint arXiv:2406.11507}, 2024.

\bibitem{liu2024unsupervised}
J.~Liu, K.~Wu, Q.~Nie, Y.~Chen, B.-B. Gao, Y.~Liu, J.~Wang, C.~Wang, and F.~Zheng, ``Unsupervised continual anomaly detection with contrastively-learned prompt,'' in \emph{Proceedings of the AAAI Conference on Artificial Intelligence}, vol.~38, no.~4, 2024, pp. 3639--3647.

\bibitem{chaudhry2018riemannian}
A.~Chaudhry, P.~K. Dokania, T.~Ajanthan, and P.~H. Torr, ``Riemannian walk for incremental learning: Understanding forgetting and intransigence,'' in \emph{Proceedings of the European conference on computer vision (ECCV)}, 2018, pp. 532--547.

\bibitem{deng2009imagenet}
J.~Deng, W.~Dong, R.~Socher, L.-J. Li, K.~Li, and L.~Fei-Fei, ``Imagenet: A large-scale hierarchical image database,'' in \emph{2009 IEEE conference on computer vision and pattern recognition}.\hskip 1em plus 0.5em minus 0.4em\relax Ieee, 2009, pp. 248--255.

\bibitem{dehaene2020anomaly}
D.~Dehaene and P.~Eline, ``Anomaly localization by modeling perceptual features,'' \emph{arXiv preprint arXiv:2008.05369}, 2020.

\bibitem{cohen2020sub}
N.~Cohen and Y.~Hoshen, ``Sub-image anomaly detection with deep pyramid correspondences,'' \emph{arXiv preprint arXiv:2005.02357}, 2020.

\bibitem{defard2021padim}
T.~Defard, A.~Setkov, A.~Loesch, and R.~Audigier, ``Padim: a patch distribution modeling framework for anomaly detection and localization,'' in \emph{International Conference on Pattern Recognition}.\hskip 1em plus 0.5em minus 0.4em\relax Springer, 2021, pp. 475--489.

\bibitem{yu2021fastflow}
J.~Yu, Y.~Zheng, X.~Wang, W.~Li, Y.~Wu, R.~Zhao, and L.~Wu, ``Fastflow: Unsupervised anomaly detection and localization via 2d normalizing flows,'' \emph{arXiv preprint arXiv:2111.07677}, 2021.

\bibitem{hyun2024reconpatch}
J.~Hyun, S.~Kim, G.~Jeon, S.~H. Kim, K.~Bae, and B.~J. Kang, ``Reconpatch: Contrastive patch representation learning for industrial anomaly detection,'' in \emph{Proceedings of the IEEE/CVF Winter Conference on Applications of Computer Vision}, 2024, pp. 2052--2061.

\bibitem{kingma2014adam}
D.~P. Kingma, ``Adam: A method for stochastic optimization,'' \emph{arXiv preprint arXiv:1412.6980}, 2014.

\bibitem{wang2022dualprompt}
Z.~Wang, Z.~Zhang, S.~Ebrahimi, R.~Sun, H.~Zhang, C.-Y. Lee, X.~Ren, G.~Su, V.~Perot, J.~Dy \emph{et~al.}, ``Dualprompt: Complementary prompting for rehearsal-free continual learning,'' in \emph{European Conference on Computer Vision}.\hskip 1em plus 0.5em minus 0.4em\relax Springer, 2022, pp. 631--648.

\bibitem{wang2022learning}
Z.~Wang, Z.~Zhang, C.-Y. Lee, H.~Zhang, R.~Sun, X.~Ren, G.~Su, V.~Perot, J.~Dy, and T.~Pfister, ``Learning to prompt for continual learning,'' in \emph{Proceedings of the IEEE/CVF conference on computer vision and pattern recognition}, 2022, pp. 139--149.

\bibitem{jia2022visual}
M.~Jia, L.~Tang, B.-C. Chen, C.~Cardie, S.~Belongie, B.~Hariharan, and S.-N. Lim, ``Visual prompt tuning,'' in \emph{European Conference on Computer Vision}.\hskip 1em plus 0.5em minus 0.4em\relax Springer, 2022, pp. 709--727.

\bibitem{eldar1997farthest}
Y.~Eldar, M.~Lindenbaum, M.~Porat, and Y.~Y. Zeevi, ``The farthest point strategy for progressive image sampling,'' \emph{IEEE transactions on image processing}, vol.~6, no.~9, pp. 1305--1315, 1997.

\bibitem{zhou2024class}
D.-W. Zhou, Q.-W. Wang, Z.-H. Qi, H.-J. Ye, D.-C. Zhan, and Z.~Liu, ``Class-incremental learning: A survey,'' \emph{IEEE Transactions on Pattern Analysis and Machine Intelligence}, 2024.

\bibitem{lin2024survey}
Y.~Lin, Y.~Chang, X.~Tong, J.~Yu, A.~Liotta, G.~Huang, W.~Song, D.~Zeng, Z.~Wu, Y.~Wang \emph{et~al.}, ``A survey on rgb, 3d, and multimodal approaches for unsupervised industrial anomaly detection,'' \emph{arXiv preprint arXiv:2410.21982}, 2024.

\bibitem{yu2024recent}
D.~Yu, X.~Zhang, Y.~Chen, A.~Liu, Y.~Zhang, P.~S. Yu, and I.~King, ``Recent advances of multimodal continual learning: A comprehensive survey,'' \emph{arXiv preprint arXiv:2410.05352}, 2024.

\end{thebibliography}

\end{document}